\documentclass{article}

\usepackage[final]{nips_2017}

\usepackage[utf8]{inputenc} 
\usepackage[T1]{fontenc}    
\usepackage{hyperref}       
\usepackage{url}            
\usepackage{booktabs}       
\usepackage{amsfonts}       
\usepackage{nicefrac}       
\usepackage{microtype}      
\usepackage{graphics} 
\usepackage{epsfig} 
\usepackage{amsmath} 
\usepackage{amssymb}  
\usepackage{multirow}
\usepackage[draft]{todonotes} 
\usepackage{color}
\usepackage{verbatim}
\usepackage{floatrow}

\newfloatcommand{capbtabbox}{table}[][1\FBwidth]

\title{Physiological and  behavioral profiling for nociceptive pain estimation using personalized multitask learning}
\author{
  Daniel Lopez-Martinez$^{1,2}$, Ognjen (Oggi) Rudovic$^2$, Rosalind Picard$^2$ \\
  $^1$ Harvard-MIT Health Sciences and Technology, \texttt{dlmocdm@mit.edu}\\
  $^2$ Affective Computing group, MIT Media Lab, Massachusetts Institute of Technology
}

\begin{document}

\maketitle

\begin{abstract}
Pain is a subjective experience commonly measured through patient's self report. While there exist numerous situations in which automatic pain estimation methods may be preferred, inter-subject variability in physiological and behavioral pain responses has hindered the development of such methods. In this work, we address this problem by introducing a novel personalized multitask machine learning method for pain estimation based on individual physiological and behavioral pain response profiles, and show its advantages in a dataset containing multimodal responses to nociceptive heat pain. 
\end{abstract}

\section{Introduction}

Pain is an unpleasant sensory and emotional experience associated with actual or potential tissue damage with sensory, emotional, cognitive and social components \cite{DeCWilliams2016}. In the clinical and research settings, pain intensity is measured using patient's self-reported pain rating scales such as the visual analog scale (VAS) \cite{Younger2009}. Unfortunately, these self-report measures only work when the subject is sufficiently alert and cooperative, and hence they lack utility in multiple situations (e.g. during drowsiness) and patient populations (e.g. patients with dementia or paralysis). 

To circumvent these limitations, automatic methods for pain estimation based on physiological autonomic signals \cite{dlm_MTL_2017,Treister2012} and/or facial expressions \cite{LopezMartinez2017c, Liu2017} have been proposed. However, inter-subject variability in pain responses has limited the ability for the automated methods to generalize across people.
For example, autonomic responses captured in  signals such as heart rate and skin conductance have been found to be  correlated only moderately with self-reported pain \cite{Dube2009}. 
A large part of the variance may be explained by inter-individual variability in autonomic reactivity, independent from stimulus intensity, and also by inter-individual variability in brain activity within structures involved in regulation of nociceptive autonomic responses \cite{Dube2009}.
Similarly, facial responses \cite{Prkachin2009} also vary across individuals 
\cite{Prkachin2008,Sullivan2000}: Some people show strong facial expressions for very low pain intensities,
while others show little or no expressiveness. Therefore, it is important to account for individual differences.

While several recent works have shown the advantages of personalization for pain estimation,  both from physiological signals \cite{dlm_MTL_2017} and  from face images \cite{Liu2017,LopezMartinez2017c}, none of these approaches explored the effect on  pain estimation performance of clustering subjects into different clusters or \textit{profiles}. Hence, in this work we investigate the grouping of subjects into different profiles based on their unique multimodal physiological and behavioral (facial expression) responses to pain. These profiles are then used to define the structure of a multi-task neural network (MT-NN), where each task in the MT-NN corresponds to a distinct \ profile. 
The advantages of this work are two-fold: (i) we show that the proposed multimodal approach achieves better performance than single-modality approaches, and (ii) we also show that by accounting for the different profiles using multi-task machine learning  we achieve further improvements in  pain estimation compared to single-task (population level) models.


\section{Dataset, Methods and Experiments} 
\label{sec:features}
We used the publicly available BioVid Heat Pain database~\cite{Walter2013}, which contains 87 participants with balanced gender ratio. The experimental setup consisted of a thermode that was used for pain elicitation on the right arm. Before the data recording started, each subject's individual \textit{pain threshold} (the temperature for which the participant's sensing changes to pain) and  \textit{tolerance threshold} (the temperature at which the pain becomes unacceptable) were determined. These thresholds were used as the temperatures for the lowest and highest pain levels together with two additional intermediate levels, thus obtaining four equally distributed pain levels ($P=\{ 1,2,3,4 \}$). For each subject, pain stimulation was applied 20 times for each of the 4 calibrated temperatures, for 4 seconds followed by a recovery phase (8-12 seconds, randomized). Each recording lasted 25 min on average. The following signals were recorded: (1) skin conductance, (2) electrocardiogram, and (3) face videos. 


\subsection{Signal processing and feature extraction} \label{sec:features}

Our pain estimation approach is posed as a continuous regression problem. Hence, for each recording we sample windows of duration 6 seconds with a step size of 0.5 seconds, and assign the label corresponding to the heat applied at the beginning of the window. To build the profiles, we used windows of duration 8 seconds, as described in Sec. \ref{sec:profiling}. In what follows, we describe the features extracted for each window.

\textbf{Skin conductance (SC): }
We used nonnegative deconvolution \cite{ledalab_dda} to decompose the SC signal into its tonic and phasic components, and to extract the phasic driver, a correlate of the activity of the sudomotor nerves \cite{Alexander2005}, such that 
$
\text{SC} = \text{SC}_\text{tonic} + \text{SC}_\text{phasic} = \text{SC}_\text{tonic} + \text{Driver}_\text{phasic} * \text{IRF}
\label{eq:decomposition}
$,  
where IRF is the impulse response function of the SC response (SCR). Then, for each window, we extracted the following features: (1) the number of SCRs with onset in the window, (2) the sum of amplitudes of all reconvolved SCRs ($\text{SC}_\text{phasic}$) with onset in the window, (3) average phasic driver activity, (4) maximum phasic driver activity, (5) integrated phasic driver activity, and (6) mean tonic activity.

\textbf{Electrocardiogram (ECG): }
From the ECG, we extracted the R peaks \cite{Engelse1979} and subsequently calculated the inter-beat intervals (IBIs). Based on previous work \cite{Werner2014a}, the following features were extracted for each window: (1) the mean of the IBIs, (2) the root mean square of the successive differences, (3) the mean of the standard deviations of the IBIs, and (4) the slope of the linear regression of IBIs in its time series. 

\textbf{Face: }
We used OpenFace \cite{openface}, an open source facial behaviour analysis toolkit, to extract for each frame the following features from the video sequences: locations of 68 facial landmarks, 3D head pose and eye gaze direction~\cite{Wood2015}, and 17 facial action units (AUs) \cite{Ekman2002a}. From the facial landmark locations, we extracted a set of geometric-based features. To do so, we first registered all landmarks using the affine transform with 4 reference points: the two eye centers, the nose center and the mouth center. Then, as in \cite{Kachele2017}, at the frame level, we calculated the euclidean distance between each of the 41 facial landmarks (we excluded the face contour and eyebrows) and the center of gravity of the facial landmark set. Therefore, each frame was represented by a 41D vector. Then, at the window level, we calculated  4 statistical features for each distance (mean, standard deviation, max, min), hence obtaining 164D geometric-based features for each window. We also calculated the same statistical descriptors for the eye gaze coordinates (24D) head pose (24D), and intensities of 17 actions units (AUs) detected by OpenFace, which are given on a scale from 0-5. 



\subsection{Subject profiling with spectral clustering} \label{sec:profiling}


To build the profiles, we extracted windows corresponding to the first 48 heat stimuli. Specifically, we used windows of duration of 8 sec., starting immediately after the onset of the stimulus. To profile the subjects based on their physiology, we calculated the SC and ECG features described in Sec. \ref{sec:features}. For facial behavioral profiles, we computed the facial expressiveness of the subjects, defined as the amount of variability in the landmark coordinates within the 8 sec. windows. From each sequence, we calculated average coordinates of each of the registered facial landmarks $l_i =(x_i,y_i)$, such that  
$
\bar{l}_i = (\bar{x}_i,\bar{y}_i) = \left ( \sum_{f=1}^F x_{f,i}, \sum_{f=1}^F y_{f,i} \right )
$, 
where $F$ is the number of frames in each 25 min sequence, and $i \in [1,68]$, provided by OpenFace~\cite{openface}. Then, for each landmark $i$ and each frame $f=1,\dots,N$, in a given window, we calculated the distance from the landmark coordinates to the mean coordinates, i.e., ${d}_{i,f} = ||{l}_{i,f} - \bar{{l}}_i||$. Finally, for each window, the facial expressiveness score is obtained as the level of variability in facial landmarks, computed as $l_{\textnormal{exp}} = \frac{1}{N} \sum_{f=1}^{N} \left ( \sum_{i=1}^{68} d_{i,f} \right )$.

\begin{figure*}
	\centering
	\includegraphics[width=1\linewidth]{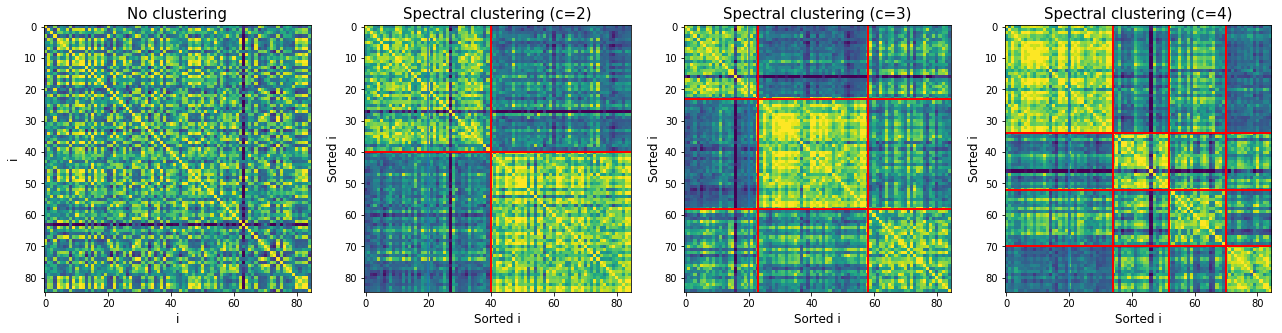}
	\caption{Subject profiling using spectral clustering. The $i,j$ elements in the (clustered) similarity matrices represent the distance from person $i$ to person $j$ in their physiological and behavioral features. The more yellow the matrix element, the closer the corresponding subjects are.}
	\label{fig:clustering}
\end{figure*}

\begin{figure}
\begin{floatrow}
\capbtabbox{%
	\scriptsize
  \begin{tabular}{|c|c|c|c|c|c|} 
\hline
\textbf{\# clusters}   & \textbf{Cluster} & \textbf{Size} & \textbf{Males} & \textbf{Females} & \textbf{Age (mean(std))}\\ \hline \hline
\multirow{1}{*}{NC} & 1  &   85     &   50.59\%    &   49.41\%      & 41.20(14.36)   \\ \hline \hline

\multirow{2}{*}{c=2} & 1   &  40     &   55.00\%    &  45.00\%       & ~44.53(13.11)$^*$    \\ \cline{2-6} 
                     & 2   &  45     &    46.67\%   &  53.33\%        & ~38.24(14.78)$^*$    \\ \hline \hline
                     
\multirow{3}{*}{c=3} & 1    &  23    &  56.52\% &  43.48\%     & ~47.61(11.24)$^*$           \\ \cline{2-6} 
                     & 2 &   35      & 48.57\%      &  51.43\%       & 38.03(14.80)    \\ \cline{2-6} 
                     & 3   &   27    &  48.15\%     &    51.85\%     & 39.85(14.42)    \\ \hline
\end{tabular}
}{%
  \caption{Cluster statistics for the clusters shown in Fig. \ref{fig:clustering}. For each cluster, statistical significance ($\textnormal{p-value} \le 0.5$) in terms of age composition with respect to the other clusters is indicated with $\ast$.}%
  \label{tab:clusters}
}
\ffigbox{%
  \centering
  \includegraphics[width=0.55\linewidth]{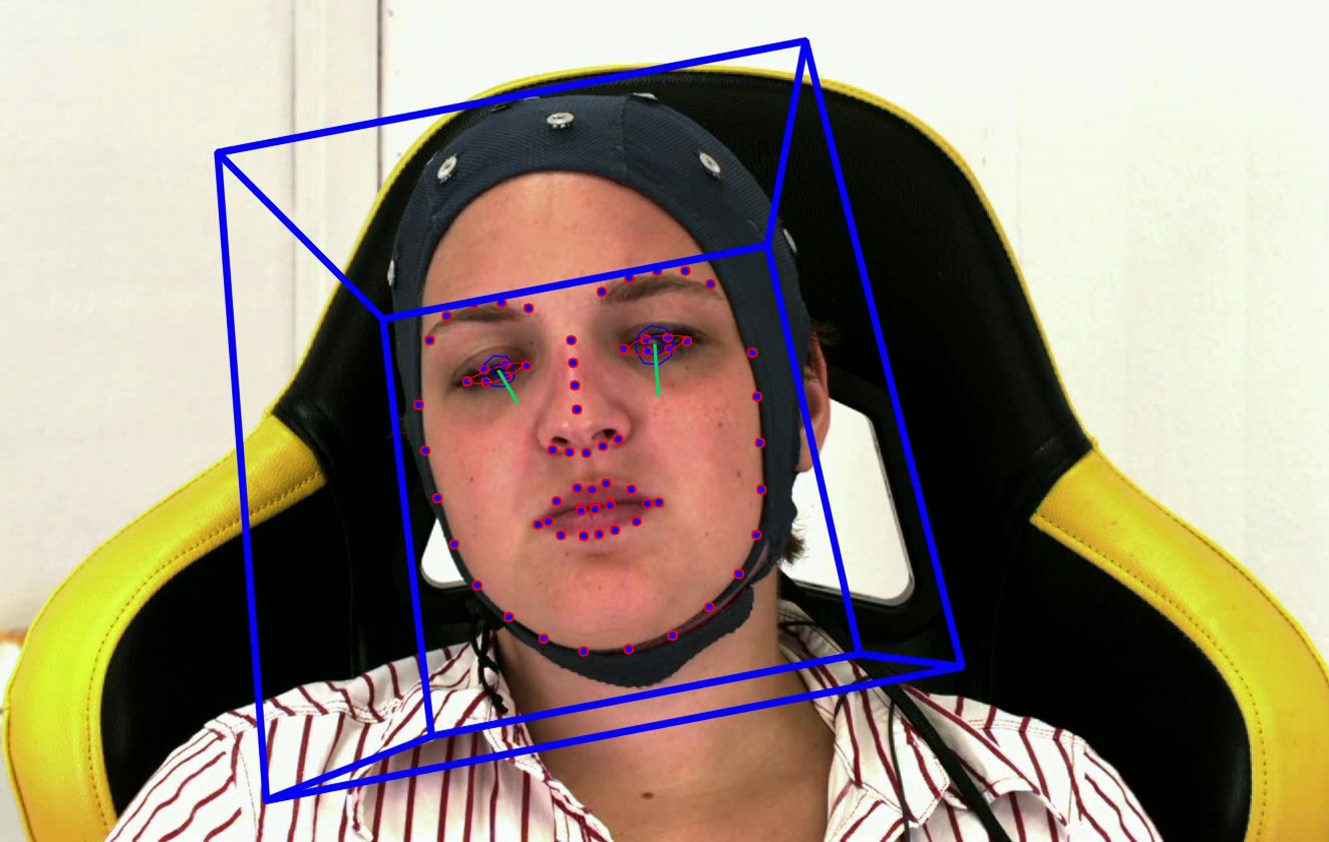} ~~~ ~~~~~ ~~~~~~~%
}{%
  \caption{Example of face tracked with\newline
    \hspace{\linewidth}OpenFace \cite{openface} showing landmarks,\newline
    \hspace{\linewidth}head pose and eye gaze.} %
}
\end{floatrow}
\end{figure}

To obtain the subject descriptors, for each subject we computed the mean value of each of the above 11 features, within windows of each of the 4 pain levels separately. Hence, each subject $s$ was represented by a 44D vector $p_s =[p_{s,1},...,p_{s,44}]$. We normalized these vectors to sum to 1 across the pain levels per subject (to compensate for the scale differences among subjects), and used them to cluster subjects into different \textit{profiles}. To this end, we used normalized spectral clustering~\cite{ShiMalik2000, Planck2006}. Namely, let $\hat{p}_s$ be the normalized descriptor of subject $s$. First, we construct a fully connected similarity graph and the corresponding weighted adjacency matrix $W$. We used the fully-conected graph, with edge weights $w_{ij}=K_{i,j} = K(p_i,p_j)$, and the radial basis function (RBF) kernel: $K(p_i,p_j) = \exp \left (-\gamma || p_i - p_j ||^2 \right )$ with $\gamma = 0.18$, as the similarity measure. 
Then, we build the degree matrix $D$ as the diagonal matrix with degrees $d_1,...,d_n$ on the diagonal, where $d_i$ is given by
$
d_i = \sum_{j=1}^{M} w_{ij}
$
where $M$ is the number of subjects in our dataset. Next, we compute the normalized graph Laplacian $L = I - D^{-1}W$ and calculate the first $c$ eigenvectors $u_1,...,u_c$ of $L$, where $c$ is the desired number of clusters. Let $U \in \mathbb{R}^{M\times c}$ be the matrix containing the vectors $u_1,...,u_c$ as columns. For $i=1,...,n$, let $y_i \in \mathbb{R}^c$ be the vector corresponding to the $i$-th row of $U$. We cluster the points $(y_i)_{y=1,...,M}$ in $\mathbb{R}^c$ with the $k$-means algorithm into clusters $C_1,...,C_c$, where $c$ was determined by visual inspection of the grouped elements of $W$ after  clustering (see Fig.\ref{fig:clustering}).


\subsection{Personalized multi-task neural network for pain level estimation}

As in \cite{dlm_MTL_2017}, we use a multi-task neural network (MT-NN) approach with shared layers and task-specific layers. While in \cite{dlm_MTL_2017} each task corresponded to a different person, here we assign a task to each profile. 
The benefits of using  profiles as  tasks are two-fold: (i) more data is available to tune the models and avoid over-fitting, and (2) in real-world applications, when a new subject arrives, only a small amount of data will need to be acquired to assign the subject to a profile, without the need to train a new layer in the MT-NN.
Specifically, in this work our regression MT-NN consisted of one shared hidden layer, and one task-specific layer. For all units in the MT-NN, we employed the rectified linear unit (ReLU) activation function: $\textbf{x}_i = \max(0,\textbf{M}_i \textbf{x}_i + \textbf{b}_i)$, where $\textbf{x}_i$ represents the input of the $i$-th layer of the network ($\textbf{x}_0$ is the input feature vector), and $\textbf{M}_i$ and $\textbf{b}_i$ are the weight matrix and bias term.
The proposed model was implemented using deep learning frameworks TensorFlow 1.2.1~\cite{Abadi2005} and Keras 2.0.6~\cite{chollet2015keras}. To optimize the network parameters, we first trained a joint network ($c$=1) using the \textit{Adam} optimizer \cite{Kingma2015}, with mean absolute error loss. We then fixed the weights of the shared layer, and used the learned weights of layer $i=2$ to initialize the profile-specific layers, which were further fine-tuned using the data corresponding to the subjects assigned to each profile. To regularize the network, we applied dropout \cite{Srivastava2014} and employed an early stopping strategy based on a validation set.

\begin{table}
\floatbox[{\capbeside\thisfloatsetup{capbesideposition={right,top},capbesidewidth=5cm}}]{table}[\FBwidth]
{\caption{Model performance in terms of mean absolute error (MAE), root-mean-square error (RMSE) and intra-class correlation (ICC) using different modalities and clustering approaches (NC for no clustering (c=1), and c=\{2,3,4\} for the clustered MT-NNs). Two-tail t-tests were  performed on the {"all"} conditions against {"multimodal (NC)"} and $\ast$ indicate significance with $\textnormal{p-value}\le 0.05$.  
} \label{tab:results}}
{\footnotesize
  \begin{tabular}{|l|l|c|c|c|}
\hline
\multicolumn{2}{|c|}{\textbf{Model}}          & \multicolumn{1}{c|}{\textbf{MAE}} & \multicolumn{1}{c|}{\textbf{RMSE}} & \multicolumn{1}{c|}{\textbf{ICC(3,1)}} \\ \hline \hline
\multicolumn{2}{|l|}{Kächele et al. \cite{Kachele2017}}          & 0.99                              & 1.16                               & N/A                               \\ \hline \hline
\multicolumn{2}{|l|}{Physiology (NC)}          & 1.00                              & 1.28                               & 0.28                                 \\ \hline
\multicolumn{2}{|l|}{Video (NC)}               & 0.92                              & 1.23                               & 0.21                                   \\ \hline
\multicolumn{2}{|l|}{Multimodal (NC)}          & 0.88                              & 1.21                               & 0.29                                  \\ \hline
\multirow{3}{*}{Multimodal (c=2)} & Cluster 1 & 0.86                              & 1.22                               & 0.22                              \\ \cline{2-5} 
                                  & Cluster 2 & 0.77                              & 1.16                               & 0.37                                 \\ \cline{2-5} 
                                  & All       & 0.82                             & 1.19                               & 0.30                                \\ \hline
\multirow{4}{*}{Multimodal (c=3)} & Cluster 1 & 0.86                              & 1.22                               & 0.19                               \\ \cline{2-5} 
                                  & Cluster 2 & 0.73                              & 1.11                               & 0.42                                 \\ \cline{2-5} 
                                  & Cluster 3 & 0.84                              & 1.19                               & 0.27                                  \\ \cline{2-5} 
                                  & All       & ~{0.80}$^\ast$                              & ~{1.17}$^\ast$                              & {\bf0.32}                             \\ \hline
\multirow{1}{*}{Multimodal (c=4)} & All & ~{\bf 0.77}$^\ast$  & ~{\bf 1.15}$^\ast$        & 0.31 \\ \hline               \end{tabular}}
\end{table}

\section{Results}
For each recording, we used the first part, corresponding to the first 48 pain stimuli, as training set, the second part corresponding to the following 10 stimuli as validation set, and the final part corresponding to the final 22 stimuli as test set. The training set was used to cluster each of the 85 subjects (2 subjects were excluded due to poor landmark tracking by OpenFace) into  $c$ clusters or \textit{profiles}. Several meta-information statistics of these clusters were calculated. They are shown in Table \ref{tab:clusters} and indicate that our clustering process does not result in any differences in the gender composition of the clusters, but it discriminates according to the subject's age. 

Once the cluster assignment was completed, we extracted features from overlapping windows as described in Sec.\ref{sec:features} We then balanced the training set to have equal amount of $P=0$ and $P>0$ instances by downsampling the over-represented class ($P=0$). As performance measures, we used  the mean absolute error (MAE),  root-mean-square error (RMSE), and  intra-class correlation ICC(3,1) \cite{Shrout1979}.
The results are shown in Table \ref{tab:results}, and also compared to related work \cite{Kachele2017}, which used a multimodal early fusion of geometric and appearance based features, SC, ECG, and electromyography (EMG). Our results indicate that similar performance is achieved with the proposed multi-modal (unclustered) approach and the prior work. The results also indicate an overall improvement of the multimodal approach with respect to the single-modality approaches, and also by the models with the proposed clustered multi-task approach. Furthermore, we note that the profiling approach also finds clusters in which pain estimation performance is better, whereas other clusters seem to be more challenging. This can be seen from the scores of cluster 3 in the  MT-NN with $c=3$,  where ICC reaches $42\%$ and achieves the best MAE and RMSE errors, thus indicating that the MT-NN model achieves improved estimation performance on this subgroup. However, it drops in performance on the other two clusters, which shows the need to further investigate the data of those subjects. By comparing the models with different number $c$ of profiles, on average we obtain similar performance, with $c=4$ performing the best in terms of MAE and RMSE, while $c=3$ attains the best ICC score.

\section{Conclusions}
We  proposed a clustered multi-task neural network model for continuous pain intensity estimation from video and physiological signals. Each task in our model represents a cluster of subjects with similar pain response profiles. We showed the benefit of our multimodal multi-task model with respect to (a) single-task (population) models, and (b) single-modality approaches. 
We conclude that the choice of  cluster profiles would need to be selected based on target application and output metric. Future work will focus on improving the profiling approach and optimizing the network topology for exploiting their commonalities.


\newpage
{\small
\bibliographystyle{unsrt}
\bibliography{Mendeley.bib}
}

\end{document}